\def\year{2021}\relax
\pdfoutput=1
\documentclass[letterpaper]{article} 
\usepackage{aaai21}  
\usepackage{times}  
\usepackage{helvet} 
\usepackage{courier}  
\usepackage[hyphens]{url}  
\usepackage{graphicx} 
\usepackage{natbib}  
\usepackage{caption} 
\usepackage[switch]{lineno} 

\usepackage{booktabs}       
\usepackage{amsfonts}       
\usepackage{amsmath,cancel}
\usepackage{subfig}
\usepackage[algo2e,noend]{algorithm2e} 
\usepackage{algorithm}
\usepackage{amsthm,bm}
\usepackage{bbm}
\newcommand{\tc}{\mathcal{C}}
\def\1{\mathbbm{1}}
\DeclareMathOperator*{\argmax}{arg\,max}
\DeclareMathOperator*{\argmin}{arg\,min}
\newcommand{\ie}{\emph{i.e.}}
\newcommand{\eg}{\emph{e.g.}}
\newcommand{\st}{\emph{s.t.}}
\newcommand{\wrt}{\emph{w.r.t.}}
\newcommand{\figref}[1]{Fig. \ref{#1}}
\newcommand{\W}{\mathcal{W}}
\newcommand{\bS}{\bar{S}}
\newcommand{\av}[1]{\langle #1 \rangle}
\usepackage[colorinlistoftodos, textwidth=26mm, shadow,color=blue!30!white]{todonotes}

\title{Fairness with Continuous Optimal Transport}
\author{
  Silvia Chiappa,\textsuperscript{\rm 1}\thanks{Equal contribution.} 
  Aldo Pacchiano\textsuperscript{\rm 2}\footnotemark[1]\\
}
\affiliations{
    \textsuperscript{\rm 1}DeepMind,
    \textsuperscript{\rm 2}UC Berkeley\\
    csilvia@google.com,
    pacchiano@berkeley.edu\\
}

\begin{document}
\maketitle
\thispagestyle{plain} 
\pagestyle{plain}

\begin{abstract}
Whilst optimal transport (OT) is increasingly being recognized as a powerful and flexible approach for dealing with fairness issues, current OT fairness methods are confined to the use of discrete OT. In this paper, we leverage recent advances from the OT literature to introduce a stochastic-gradient fairness method based on a dual formulation  of  continuous OT. We show that this method gives superior performance to discrete OT  methods when little data is available to solve the OT problem, and similar performance otherwise. We also show that both continuous and discrete OT methods are able to continually adjust the model parameters to adapt to different levels of unfairness that might occur in real-world applications of ML systems. 
\end{abstract}
\section{Introduction}
ML fairness techniques aim at ensuring that models do not encode nor amplify human and societal biases contained in the training data, and therefore that individuals are not treated unfavorably on the basis of race, gender, disabilities, or other \emph{sensitive attributes} \cite{gajane2017formalizing,mitchell2018fair,verma2018fairness,barocas2019fairness,oneto2020fairness}. 

Popular fairness criteria such as \emph{demographic parity} require certain characteristics of relevant distributions to be invariant to different sensitive attributes.
Whilst facilitating model evaluation and design, in many real-world scenarios invariance of the entire distributions might instead be necessary to ensure fairness. 

\emph{Optimal transport} (OT) \cite{villani2009optimal,peyre2019computational} is increasingly being explored to impose statistical independence \wrt~sensitive attributes through enforcing small OT distances between distributions, allowing to achieve a stronger version of demographic parity as well as other criteria \cite{black2019fliptest,delbarrio2019obtaining, johndrow2019algorithm,risser2019using,wang2019repairing,chzhen2020fair,delbarrio2020review,legouic2020price,yurochkin2020training}.

As observed in \citet{jiang2019wasserstein,chiappa2020general}, OT distances can explicitly be related to expected changes in model accuracy. This makes the application of OT to fairness particularly attractive, as it allows to devise methods for achieving statistical independence with minimal loss of accuracy and therefore with optimal fairness-accuracy trade-offs. 

Current methods approximate the computation of OT distances between continuous distributions with discrete OT. Building on the recent advances in avoiding such approximations \cite{genevay2016stochastic,genevay2019entropy,mensch2020online}, we introduce a stochastic-gradient fairness method based on a dual formulation of continuous OT. We show that this method gives superior performance to discrete OT methods when little data is available to solve the OT problem, and similar performance otherwise. 
In addition, we show that OT methods are able to continually adjust the model parameters to adapt to changes in level of unfairness,
and therefore demonstrate their suitability in real-world applications in which the data used for system training and deployment differ in unfairness levels.
\section{Related Work}
Recent literature on fairness has seen an increase in the adoption of OT. Our approach is close to the penalty approach proposed in \citet{jiang2019wasserstein}, and differs from it in using continuous, rather than discrete, OT. Whilst detailed and evaluated on classification and on \emph{strong demographic parity}, in a similar way as done in \citet{chiappa2020general} our method can be modified to solve regression problems and to achieve other fairness criteria, including strong equalized odds and causal criteria such as (path-specific) counterfactual fairness. Unlike our method, other in-processing methods to achieve strong demographic parity such as adversarial ones focus on binary sensitive attributes \cite{zhang2018mitigating,celis2019improved}. 

The adjustment capabilities to changing levels of unfairness that we show in this paper can be placed within the broader context of recent efforts in the ML fairness literature to go beyond the static fairness assumption. However, these capabilities are complementary to those of approaches accounting for downstream impact of decisions \cite{joseph2016fairness,jabbari2017fairness,blum2018preserving,gillen2018online,liu2018delayed,creager2019causal,sun2019learning,tabibian2019consequential,wen2019fairness, zhang2020fairness}. 
These latter approaches aim at prospectively addressing long-term effects that induce changes in unfairness levels during the training of the model. We instead aim at retrospectively modify the model parameters to account for different unfairness levels.
Such differences could be due to downstream impact of decisions or to other, social, factors, \eg, for the case of a risk assessment instruments, to differing rates of policing in neighborhoods that are more disproportionately populated by non-Whites  \cite{lum2016predict,rosenberg2018immigration}.
These adjustment capabilities are also different from robustness to changes in the data distribution enforced during training of the model \cite{mandal2020ensuring}.
\section{SDP-Fair Classification with OT}\label{sec:sdpot}
We focus on the problem of enforcing strong demographic parity (SDP) in binary classification.   

Consider the problem of learning a binary classification model from a dataset $\mathcal{D} =\{(a^n,x^n,y^n)\}_{n=1}^N$---with underlying joint distribution $p(A,X,Y)$---corresponding to $N$ individuals, in which $x^n \in \mathbb{R}^d$ is a vector of features, $a^n\in \mathcal{A}=\mathbb{N}^k$ is a vector of attributes that are considered sensitives, and $y^n$ is a binary class label. 

We assume that the model output for individual $n$, $s^n$, represents an estimate of the probability of belonging to class 1, $p(Y=1|A=a^n,X=x^n)$. A prediction $\hat y^n$ of $y^n$ is then obtained as $\hat y^n=\1_{s^n>\tau}$, where $\1_{s^n>\tau}=1 \text{ if } s^n>\tau$ for a threshold $\tau\in[0,1]$ and zero otherwise. The model outputs induce a random variable $S$. We indicate with $p_S$ the distribution of $S$ and with $p_{S_a}$ the conditional distribution of $S$ given sensitive attributes $a$.

SDP requires statistical independence between $S$ and $A$, and can be expressed as
\begin{align*}
\textrm{SDP: } ~~~p_{S_a} = p_{S_{\bar a}}, \hskip0.1cm \forall a,\bar a\in\mathcal{A}\,.
\end{align*}
Over demographic parity $\big(\av{\hat Y}_{p(\hat Y|A=a)} = \av{\hat Y}_{p(\hat Y|A=\bar a)}\big)$, SDP has the advantage of ensuring that the class predictions do not depend on the sensitive attributes regardless of the value of the threshold $\tau$ used. SDP is also  desirable if scores rather than binary predictions are the target outcomes---such as \eg~in risk assessment instruments where scores represent risk of re-committing a crime---as SDP would ensure that the scores do not depend on the sensitive attributes.

Our approach to impose SDP follows closely the penalty approach introduced in \citet{jiang2019wasserstein}. After training the model using a standard loss, we modify the obtained model parameters to minimize the OT distances between the groups distributions $\{p_{S_a}\}$ and a target distribution $p_{\bS}$. However, we propose a method that does not require approximating such distances with discrete OT. Whilst any target distribution $p_{\bS}$ and type of OT distance would allow to achieve SDP, for classification using the barycenter of $\{p_{S_a}\}$ and the Wasserstein-1 distance ensures optimal fairness-accuracy trade-offs (see \citet{jiang2019wasserstein}). 
\section{Computation of OT Distances}\label{sec:cot}
As part of a recent effort toward providing solutions for the computation of OT distances between continuous distributions, \citet{genevay2016stochastic,genevay2019entropy} introduced a stochastic-gradient method for solving an entropy regularized continuous OT problem using samples from the distributions. We propose to compute OT distances between $\{p_{S_a}\}$ and $p_{\bS}$ by adapting this method to our setting.
\subsection{Background on Continuous OT} 
Let $p_{X}$ and $p_{Y}$ be two probability density functions (pdfs) on ${\cal X}$ and ${\cal Y}$, $\Gamma(p_X, p_Y)$ the set of joint pdfs on ${\cal X}\times {\cal Y}$ with marginals $p_X$ and $p_Y$, and $\tc:{\cal X}\times {\cal Y}\rightarrow[0,\infty]$ a \emph{cost function}. 

The Kantorovich OT problem \cite{kantorovich42on,villani2009optimal,peyre2019computational} consists in finding a $\gamma^*\in \Gamma(p_{X}, p_{Y})$, called the \emph{optimal coupling} between $p_X$ and $p_Y$, minimizing the expected cost with respect to $\gamma(x,y)$, \ie~such that
\begin{align}\label{eq:otp}
    \gamma^* = \argmin_{\gamma\in \Gamma(p_{X}, p_{Y})} \av{\tc(x,y)}_{\gamma(x,y)},
\end{align}
where $\av{\tc(x,y)}_{\gamma(x,y)}:=\int_{\mathcal{X} \times \mathcal{Y}} \tc(x, y)\gamma(x,y)dxdy$. Under appropriate conditions on $\tc$, 
$$\W_\tc(p_{X}, p_{Y}) := \min_{\gamma\in\Gamma(p_X, p_Y)} \av{\tc(x, y)}_{\gamma(x,y)}$$ 
is a distance between $p_X$ and $p_Y$, called \emph{OT distance}. 

Our setting ($S\in[0,1]$) gives $\mathcal{X} = \mathcal{Y} = [0,1]$, and we consider $\tc(x, y) = \| x - y \|_1$, where $\| \cdot \|_1$ indicates the $L^1$-norm. The resulting OT distance, that we denote with the shorthand $\W_1(p_{X}, p_{Y})$, is known as the \emph{Wasserstein-$1$ distance}.
\subsection{Dual Formulation of Regularized  Continuous OT} 
Problem (\ref{eq:otp}) has dual formulation given by 
\begin{align}\label{eq:opt_transport_dual}
&\lambda_X^*, \lambda_Y^* = \!\!\argmax_{\lambda_X \in \mathbb{C}(\mathcal{X}), \lambda_{Y} \in \mathbb{C}(\mathcal{Y})} \av{\lambda_X(x)}_{p_{X}(x)} + \av{\lambda_Y(y)}_{p_{Y}(y)}\nonumber\\
&~~~~~~~~~~~~~~~~~~~~~~~~\st~\lambda_X(x) + \lambda_Y(y) \leq \tc(x,y)~~~\forall x, y,
\end{align}
where $\mathbb{C}(\cdot)$ indicates the space of continuous functions. The dual formulation has the advantage of containing expectations \wrt~the marginals $p_X$ and $p_Y$, which can be estimated using samples from these distributions. 
However, due to the difficulty in fulfilling the constraint $\lambda_X(x) + \lambda_Y(y) \leq \tc(x,y)$, \citet{genevay2016stochastic,genevay2019entropy} suggest to consider a regularized version of Problem (\ref{eq:otp}) given by
\begin{align}\label{eq:otpr}
\gamma^* = \argmin_{\gamma\in \Gamma(p_{X}, p_{Y})} \av{\tc(x,y)}_{\gamma(x,y)} + \lambda {\cal D}_{\phi}[\gamma|p_{X}p_{Y}],
\end{align}
where ${\cal D}_{\phi}[\gamma|p_{X}p_{Y}]:=\Bigl< \phi\!\left(\frac{\gamma(x,y)}{p_X(x)p_Y(y)}\right)\Bigr>_{\!p_X(x)p_Y(y)}$
is the $\phi$-divergence between $\gamma$ and $p_Xp_Y$ \cite{csiszar1975divergence}, and $\lambda$ is a scalar that controls the regularization strength. We indicate the regularized OT distance with $\W^{\lambda}_\tc(p_{X}, p_{Y})$.

Following \citet{genevay2016stochastic,seguy2018largescale,genevay2019entropy,lee2020convergence}, we consider\footnote{We consider $\phi(x) = x\log(x) - x$ rather than $\phi(x) = x\log(x)$, as this leads to a simpler formulation.} $\phi(x)=x\log(x)-x$, obtaining the \emph{relative entropy regularization}
\begin{align*}
{\cal D}_{\phi}[\gamma|p_{X}p_{Y}]=\Biggl<\log\left(\frac{\gamma(x,y)}{p_{X}(x)p_{Y}(y)}\right)-1\Biggr>_{\!\!\gamma(x,y)},
\end{align*}
and $\phi(x)=x^2+i_{\mathbb{R}^+}(x)$, where $i$ denotes the convex indicator function, obtaining the \emph{$L^2$ regularization}
\begin{align*}
{\cal D}_{\phi}[\gamma|p_{X}p_{Y}]=\Biggl<\left(\frac{\gamma(x,y)}{p_{X}(x) p_{Y}(y)}\right)^2\Biggr>_{\!\!p_X(x)p_Y(y)}.
\end{align*}
Problem (\ref{eq:otpr}) has dual formulation given by
\begin{align}\label{eq:otpr_dual}
\lambda_X^*, \lambda_Y^*=& \argmax_{\lambda_X \in \mathbb{C}(\mathcal{X}), \lambda_{Y} \in \mathbb{C}(\mathcal{Y})} \Biggl<\lambda_X(x)+\lambda_Y(y)\Bigr.\\
&\!\!\Bigl. - \lambda\phi^*\!\left(\frac{\lambda_X(x)+\lambda_Y(y)-\tc(x,y)}{\lambda}\right)\!\Biggr>_{\!\!p_{X}(x)p_{Y}(y)}\!,\nonumber
\end{align}
where $\phi^*(x)=\exp(x)$ for the relative entropy regularization, and 
$\phi^*(x) =\max(x,0)^2/4$ for the $L^2$ regularization. 
\begin{proof}
Indicating with $\lambda_X$ and $\lambda_Y$ the Lagrange multipliers, we can write the dual function as 
\begin{small}
\begin{align*}
&\!\!\min_{\gamma(x,y)} \av{\tc(x,y)}_{\gamma(x,y)} 
+ \lambda \Bigl<\phi\!\left(\frac{\gamma(x,y)}{p_X(x)p_Y(y)}\right)\Bigr>_{p_X(x)p_Y(y)}\\
&\hskip0cm + \Bigl<p_X(x) \!-\! \int_Y\!\gamma(x,y)dy\Bigr>_{\lambda_X(x)} 
\!\!\!\!\!+ \Bigl<p_Y(y) \!-\! \int_X\!\gamma(x,y)dx\Bigr>_{\lambda_Y(y)}\\
&\!=\!\av{\lambda_X(x)}_{p_X(x)} +\av{\lambda_Y(y)}_{p_Y(y)}+\lambda\!\min_{\gamma(x,y)} \Biggl<\phi\! \left(\frac{\gamma(x,y)}{p_X(x)p_Y(y)}\right)\\
&~~~~~~~~~~~~~~~~~ -\frac{\lambda_X(x) \!+\! \lambda_Y(y) \!-\! \tc(x,y)}{\lambda}\frac{\gamma(x,y)}{p_X(x)p_Y(y)}\Biggr>_{\!\!p_X(x)p_Y(y)}\\
&\!=\!\Biggl<\!\!\lambda_X(x)\!+\!\lambda_Y(y)\!-\! \lambda\phi^*\!\left(\!\frac{\lambda_X(x)\!+\!\lambda_Y(y)\!-\!\tc(x,y)}{\lambda}\!\right)\!\!\Biggr>_{\!\!p_{X}(x)p_{Y}(y)}\!\!\!,
\end{align*}
\end{small}
\!\!where $\phi^*$ is the Legendre transform of $\phi$, \ie~$\phi^*(x^*):=\max_x (x^*x-\phi(x))$.
\end{proof}
\subsection{Dual Variables Parametrization}
Following \citet{genevay2016stochastic,genevay2019entropy}, we parametrize the \emph{dual variables} $\lambda_X$ and $\lambda_Y$ as expansions in reproducing kernel Hilbert spaces using the random Fourier features approximation introduced in \citet{rahimi2008random}. More specifically, we approximate a Gaussian kernel $\kappa$ with variance $\sigma^2$ as
\begin{align*}
\kappa(x,x')\approx\frac{2}{D}\sum_{i=1}^D \cos\big(\omega^i x+b^i\big)\cos\big(\omega^i x'+b^i\big),
\end{align*}
where $\omega^i\sim {\cal N}(0,2/\sigma^2)$ and $b^i\sim U[0, 2\pi]$, with ${\cal N}(\cdot,\cdot)$ and $U[\cdot,\cdot]$ indicating the Gaussian and uniform distributions respectively.
This way we can express $\lambda_X(x)$ as 
$$\lambda_X(x) \approx  \lambda_X^\top   \psi_{\lambda_X}(x),$$ 
where $\lambda_X$ is a $D$-dimensional vector\footnote{Throughout the paper, we use an abuse of notation by overloading the symbol $\lambda_X$ to denote both a function and a vector. The meaning should be clear from the context.} and 
$$\psi_{\lambda_X}(x)=\sqrt{\frac{2}{D}}\left(\cos\big(\omega^1 x+b^1\big),\ldots,\cos\big(\omega^Dx+b^D\big)\right)^{\!\top}\!\!.$$
It is possible to show that, if $D = \tilde{\mathcal{O}}(1/\epsilon^2)$, the random feature maps can achieve an approximation error of $\epsilon$ \wrt~the original Gaussian kernel (see Claim 1 in \citet{rahimi2008random}). Experimentally, we did not see significant difference in performance for $D\geq 50$.
\subsection{Optimal Dual Variables}
Using samples $\{x^i\}_{i=1}^{N_x}$ and $\{ y^j \}_{j=1}^{N_y}$ from $p_X$ and $p_Y$, we can form a Monte-Carlo approximation of the expectation in Problem (\ref{eq:otpr_dual}), and find the optimal dual variables through stochastic gradient.
Specifically, the gradient with respect to $\lambda_X$  is given by
\begin{align*}
&\nabla_{\lambda_X}\left(\frac{1}{N_x}\sum_{i=1}^{N_x}\lambda_X(x^i)+ \frac{1}{N_y}\sum_{j=1}^{N_y}\lambda_Y(y^j)\right.\nonumber\\
&~~~~\left.-\frac{\lambda}{N_xN_y}\sum_{i=1}^{N_x}\sum_{j=1}^{N_y}\phi^*\!\left(\frac{\!\lambda_X(x^i)+\lambda_Y(y^j)-\tc(x^i,y^j)}{\lambda}\right)\!\right)\nonumber\\
&~~~~=\frac{1}{N_x}\sum_{i=1}^{N_x}\left(1 - \frac{\lambda}{N_y}\sum_{j=1}^{N_y}\phi^*(x^i, y^j)'\right)\psi_{\lambda_X}(x^i),
\end{align*}
where $\phi^*(x^i, y^j)' = \frac{1}{\lambda}\exp\left(\frac{\lambda_X(x^i)+\lambda_Y(y^j)-\tc(x^i,y^j)}{\lambda}\right)$ for the relative entropy regularization, whilst $\phi^*(x^i, y^j)'=\frac{1}{2\lambda^2}\left(\lambda_X(x^i)+\lambda_Y(y^j)-\tc(x^i,y^j)\right)$ for the $L^2$ regularization. This suggests the following stochastic-gradient update 
\begin{align}\label{eq:update_dual}
\lambda_X \leftarrow \lambda_X + \frac{\epsilon_{\lambda}}{N_x}\sum_{i=1}^{N_x}\left(1-\frac{1}{N_y}\sum_{j=1}^{N_y}\alpha_{ij}\right)\psi_{\lambda_X}(x^i),
\end{align}
where $\epsilon_{\lambda}$ is the update size and $\alpha_{ij}:=\lambda \phi^*(x^i, y^j)'$.
\paragraph{Enforcing $\mathbf{\lambda_X(x) + \lambda_Y(x) = 0}$.} In the limit $\lambda \rightarrow 0$, for a cost function satisfying $\mathcal{C}(x,x) = 0 ~\forall x$, the soft constraint given by $ -\lambda\phi^*\!\left(\frac{\lambda_X(x)+\lambda_Y(y)-\tc(x,y)}{\lambda}\right)$ in Problem (\ref{eq:otpr_dual}) converges to the hard constraint of Problem (\ref{eq:opt_transport_dual}). Whenever the hard constraint is enforced, $\lambda_X(x) +\lambda_Y(x) \leq 0~\forall x$, and therefore $\lambda_Y(x) = -\lambda_X(x) - \xi(x)$ for some non-negative function $\xi$, giving
\begin{align*}
&\av{\lambda_X(x)}_{p_{X}(x)} + \av{\lambda_Y(y)}_{p_{Y}(y)}\\
&~~~~~~~~~~~~=\av{\lambda_X(x)}_{p_{X}(x)} - \av{\lambda_X(y)}_{p_{Y}(y)} - \av{\xi(y)}_{p_{Y}(y)}\\
&~~~~~~~~~~~~\leq \av{\lambda_X(x)}_{p_{X}(x)} - \av{\lambda_X(y)}_{p_{Y}(y)}.
\end{align*}
If $\lambda_X$ and $-\lambda_X$ also satisfy the constraint of Problem (\ref{eq:opt_transport_dual}), the duple ($\lambda_X, - \lambda_X$) is a better candidate to achieve the optimum of Problem (\ref{eq:opt_transport_dual}), suggesting that we can restrict the optimization to ($\lambda_X, \lambda_Y$) satisfying 
$$\lambda_X(x) + \lambda_Y(x) = 0.$$ 
This argument is used for example when $\mathcal{C}(x, y) = \| x \|_1$ and $\lambda_X, \lambda_Y$ are constrained to be $1-$Lipschitz \cite{arjovsky2017wasserstein}.
We use this strategy to stabilize the optimization problem. With this constraint, update (\ref{eq:update_dual}) needs to be adjusted by adding to it the term 
\begin{align*}
-\frac{\epsilon_{\lambda}}{N_y}\sum_{j=1}^{N_y}\left(1-\frac{1}{N_x}\sum_{i=1}^{N_x}\alpha_{ij}\right)\psi_{\lambda_X}(y^j).
\end{align*}
\begin{figure}[t]
\centering
\hskip-0.1cm
\includegraphics[height=2cm,width=2.9cm]{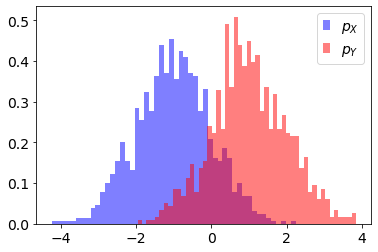} 
\includegraphics[height=2cm,width=2.7cm]{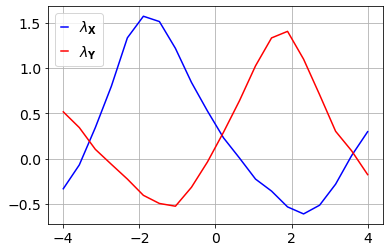}
\includegraphics[height=2cm,width=2.7cm]{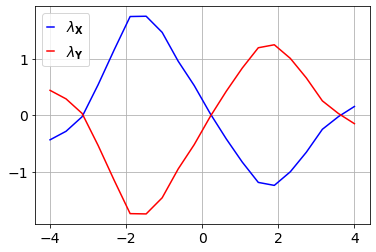}
\caption{\textbf{Left:} Histograms of two Gaussians $\mathcal{N}(-1, 1)$ and $\mathcal{N}(1, 1)$. \textbf{Center:}  Optimal dual variables obtained using the stochastic-gradient method described above. \textbf{Right:} Optimal dual variables with constraint $\lambda_X(x) +\lambda_Y(x)=0$.
}
\label{fig:dual_functions}
\end{figure}
\paragraph{Intuition on Dual Variables.} If we disregard the last term in Problem (\ref{eq:otpr_dual}), we obtain $\mathcal{W}^{\lambda}_\mathcal{C}(p_X, p_Y) \approx \av{\lambda^*_X(x)}_{p_{X}(x)} + \av{\lambda^*_Y(y)}_{p_{Y}(y)}$. This indicates that the optimal dual variable $\lambda_X^*$ achieves large values in regions of the support where $p_X$ differs the most from $p_Y$ and vice-versa, as illustrated in \figref{fig:dual_functions} for two Gaussian distributions with differing means.
\section{COT Method}\label{sec:fcot}
In this section, we introduce a stochastic-gradient method that modifies the parameters of a binary classifier in order to achieve SDP using the approach for computing OT distances described above. 

We assume that the model output $s^n$ is obtained with the logistic function, \ie
\begin{align*}
s^n=g_{\theta}(a^n, x^n):=1/\big(1 + \exp\big(-\theta^\top (a^n, x^n)^\top\big)\big),
\end{align*}
where $\theta \in \mathbb{R}^{k+d}$ are the model parameters. After training the model using the standard logistic regression loss, we modify the obtained $\theta^*$ to minimize the sum over $a \in \mathcal{A}$ of the regularized Wasserstein-1 distances between $p_{S_a}$ (which depends on $\theta$) and a target distribution $p_{\bS}$ (which we keep fixed at $\theta^*$), \ie~to solve the following problem
\begin{align}\label{eq:min_wass}
\min_{\theta}\sum_{a \in \mathcal{A}}\W^{\lambda}_{1}(p_{S_a}, p_{\bar S}).
\end{align}
Indicating with $\lambda_{S_a}$ and $\lambda_{\bar S, a}$ the dual variables associated to $\W^{\lambda}_1(p_{S_a}, p_{\bar S})$,
the dual formulation of Problem (\ref{eq:min_wass}) can be expressed as 
\begin{align}\label{eq:min_wass_dual}
&\hskip-0.2cm \min_{\theta}~\sum_{a \in \mathcal{A}}~\max_{\lambda_{S_a}, \lambda_{\bar S,a}} \left(\vphantom{\frac{1}{2}}\bigl<\lambda_{S_a}(s)\bigr>_{p_{S_a(s)}} + \bigl<\lambda_{\bar S, a}(\bar s)\bigr>_{p_{\bar S(\bar s)}}-\right.\nonumber\\
&\left.\Bigl<\lambda \phi^*\left(\frac{\lambda_{S_a}(s)+\lambda_{\bar S, a}(\bar s)-\tc(s, \bar s)}{\lambda }\right)\Bigr>_{p_{S_a(s)}p_{\bar S(\bar s)}}\right).
\end{align}
We propose to solve Problem (\ref{eq:min_wass_dual}) with a stochastic-gradient approach that alternates between the following two steps:
\begin{itemize}
\item[1.] Perform a gradient update of the dual variables $\lambda_{S_a}$ and $\lambda_{\bar S,a}$, $\forall a \in \mathcal{A}$, keeping the model parameters $\theta$ fixed. 
\item[2.] Perform a gradient update of the model parameters $\theta$, keeping the dual variables $\lambda_{S_a}$ and $\lambda_{\bar S,a}$ fixed. 
\end{itemize}
The steps are achieved with samples $\{(a^i=a,x^i,y^i)\}_{i=1}^{N_{S}}$ $\forall a \in \mathcal{A}$ from the training dataset ${\cal D}$, which are used to create the model outputs  $\{s^i_a=g_{\theta}(a,x^i)\}_{i=1}^{N_{S}}$, and with samples $\{\bar s^j\}_{j=1}^{N_{\bar S}}$ from a dataset $\bar{\mathcal{D}}$ representing the target distribution $p_{\bar S}$, and have computational cost ${\cal O}(N_{S}(N_{\bar S} + D))$.
\begin{algorithm}[t]
\textbf{Input: Regularization strength $\lambda$.
Number of random Fourier features $D$. 
Kernel variance $\sigma^2$. 
Dual variables and parameters update sizes $\epsilon_{\lambda}, \epsilon_{\theta}$. 
Number of updates $K$. Batch sizes $N_{S}, N_{\bar S}$.}\\
Learn $\theta^*$ with standard logistic regression. Set $\theta=\theta^*$.\\
Create dataset $\bar{\mathcal{D}}$ representing $p_{\bar S}$.\\
Initialize dual variables as $\lambda_{S_a} = \mathbf{0}$, $\lambda_{\bar S,a} = \mathbf{0}$, $\forall a\in\mathcal{A}$.\\
\For{$k=1, \cdots, K$ }{
Sample $\{\bar s^j_t\}_{j=1}^{N_{\bar S}}$ from $\bar{\mathcal{D}}$. \\
\For{$a\in\mathcal{A}$}{
Sample $\{(a^i=a,x^i, y^i)\}_{i=1}^{N_{S}}$ from $\mathcal{D}$ and create the model outputs $\{s^i_{a}=g_{\theta}(a,x^i)\}_{i=1}^{N_{S}}$.\\
Compute $\alpha_{ij} = \lambda \phi^*(s^i_{a},\bar s^j)'$.\\
Update dual variables\\
\begin{small}
   $\lambda_{S_a} \leftarrow \lambda_{S_a} + \frac{\epsilon_{\lambda}}{N_{S_a}}\sum_{i=1}^{N_{S}}\left(1\!-\!\frac{\sum_{j=1}^{N_{\bar S}}\alpha_{ij}}{N_{\bar S}}\right)\psi_{\lambda_{S_a}}(s^i_a)$,\\
   $\lambda_{\bar S,a} \leftarrow \lambda_{\bar S,a} + \frac{\epsilon_{\lambda}}{N_{\bar S}}\sum_{j=1}^{N_{\bar S}}\left(1\!-\!\frac{\sum_{i=1}^{N_{ S}}\alpha_{ij}}{N_{S}}\right)\psi_{\lambda_{\bar S,a}}(\bar s^j)$.
\end{small}
}
Update parameters
\begin{small}
\begin{align*} 
\theta \leftarrow \theta - \epsilon_{\theta}\sum_{a \in \mathcal{A}}\sum_{i=1}^{N_{S}}\sum_{j=1}^{N_{\bar S}}\Big(&(1-\alpha_{ij})\nabla_{g_{\theta}}\lambda_{S_a}(s^i_{a}) \\
&~~~~~+\alpha_{ij}\nabla_{g_{\theta}}\tc(s^i_{a},\bar s^j)\big)\Big)\nabla_{\theta}s^i_{a}.
\end{align*} 
\end{small}
}
\textbf{Return:} $\theta; \lambda_{S_a}, \lambda_{\bar S,a}$ $\forall a\in\mathcal{A}$.
\caption{{\bf COT Algorithm}}
\end{algorithm}
The gradient updates are explained in detail below. The full procedure is summarized in the COT Algorithm above.
\paragraph{1. Dual Variables Update.}
Using the explanation above, in the unconstrained case the updates for the dual variables are obtained as 
\begin{align*}
&\lambda_{S_a} \leftarrow \lambda_{S_a} + \frac{\epsilon_{\lambda}}{N_{S_a}}\sum_{i=1}^{N_{S}}\left(1-\frac{1}{N_{\bar S}}\sum_{j=1}^{N_{\bar S}}\alpha_{ij}\right)\psi_{\lambda_{S_a}}(s^i_a),\\
&\lambda_{\bar S,a} \leftarrow \lambda_{\bar S,a}+ \frac{\epsilon_{\lambda}}{N_{\bar S}}\sum_{j=1}^{N_{\bar S}}\left(1-\frac{1}{N_{S}}\sum_{i=1}^{N_{ S}}\alpha_{ij}\right)\psi_{\lambda_{\bar S,a}}(\bar s^j).
\end{align*}
\begin{table*}[t]
\setlength{\tabcolsep}{5pt}
\centering
\scalebox{.95}{
\begin{tabular}{lcccc|cccc|cccc}
&\multicolumn{4}{c}{Adult}&\multicolumn{4}{c}{German Credit}&\multicolumn{4}{c}{Community \& Crime}\\
\cline{2-5}\cline{6-9}\cline{10-13}
& Err-.5 & Wass1 & SDD & SPDD     & Err-.5 & Wass1 & SDD  & SPDD   & Err-.5 & Wass1 & SDD  & SPDD\\
\toprule
LR & .142 & .313 & .426 & .806     & .248 & .103 & .102 & .102    & .116 & 1.435 & 1.402 & 7.649\\
DOT & .175 & .027 & .023 & .054     & .282 & .035 & .020 & .020    & .327 &  .207 &  .176 &  .848\\
DPP & .170 & .025 & .017 & .043     & .248 & .032 & .024 & .024    & .327 &  .356 &  .159 &  .821\\
COT & .175 & .023 & .020 & .044     & .242 & .042 & .027 & .027    & .324 &  .223 &  .204 &  .928\\
\bottomrule
\end{tabular}}
\caption{Test classification error (Err-.5) and unfairness level (Wass1, SDD, and SPDD) for logistic regression (LR), the methods in \citet{jiang2019wasserstein} (DOT and DPP), and COT on the UCI Adult, German Credit, and Community \& Crime datasets.}
\label{table:AGC}
\end{table*}
\paragraph{2. Model Parameters Update.}
The first and third expectations in Problem (\ref{eq:min_wass_dual}) are \wrt~the model outputs distribution $p_{S_{a}}$ which (unlike $p_{\bar S}$) is a function of $\theta$, making the minimization \wrt~ $\theta$ a challenging task. This difficulty can be eliminated by rewriting these expectations as \wrt~the data distribution. For example, we can rewrite $\av{\lambda_{S_a}(s)}_{p_{S_{a}(s)}}$ as
\begin{align*}
\av{\lambda_{S_a}(s)}_{p_{S_{a}(s)}}=\av{\lambda_{S_a}(g_{\theta}(a,x))}_{p(x|a)}, 
\end{align*}
and then make a Monte-Carlo approximation using $\{s^i_a=g_{\theta}(a,x^i)\}_{i=1}^{N_{S}}$. Indeed, using the change of variables rule for $s=g_{\theta}(a,x)$ and 
$p_{S_a}(s)|\partial s/\partial x|=p(x|a)$, we can write
\begin{align*}
\av{\lambda_{S_a}(s)}_{p_{S_{a}(s)}}&=
\int_{x}p_{S_a}(g_{\theta}(a,x))\lambda_{S_a}(g_{\theta}(a,x))|\partial s/\partial x|\\
&=\int_{x}p(x|a)\lambda_{S_a}(g_{\theta}(a,x)).
\end{align*}
Using the chain rule, we obtain the stochastic-gradient update\footnote{With the term $\frac{1}{N_{S}N_{\bar S}}$ absorbed into $\epsilon_{\theta}$.}
\begin{align*}
\theta \leftarrow \theta - \epsilon_{\theta}\sum_{a \in \mathcal{A}}\sum_{i=1}^{N_{S}}\sum_{j=1}^{N_{\bar S}}\Big(&(1-\alpha_{ij})\nabla_{g_{\theta}}\lambda_{S_a}(s^i_{a}) \\
&~~~~~+\alpha_{ij}\nabla_{g_{\theta}}\tc(s^i_{a},\bar s^j)\big)\Big)\nabla_{\theta}s^i_{a},
\end{align*}
with $\nabla_{g_{\theta}}\lambda_{S_a}(s^i_a)=-\sqrt{\frac{2}{D}}\lambda_{S_a}^\top\left(\omega^1\sin\left(\omega^1 s^i_a+b^1\right),\ldots,\right.$ 
$\left.\omega^D\sin\left(\omega^D s^i_a+b^D\right)\right)$, $\nabla_{g_{\theta}}\tc(s^i_a,\bar s^j)=\textrm{sign}(s^i_a-\bar s^j)$, and $\nabla_{\theta}s^i_a=(a,x^i)^\top s^i_a(1-s^i_a)$.
\section{Results}
We first compare the COT method with two discrete OT methods introduced in \citet{jiang2019wasserstein} in the standard static fairness scenario.
We show superior performance of COT when little data is available to solve the OT problem, and similar performance otherwise. We then consider a scenario in which the unfairness level is assumed to change, and show that OT methods are able to continually adjust the model parameters to adapt to such changes.

The first method introduced in \citet{jiang2019wasserstein} (referred to as \emph{DOT}) can be seen as the discrete counterpart of COT. More specifically, DOT alternates between the following two steps:
\begin{itemize}
\item[1.] Estimate the optimal coupling matrix as
\begin{align}\label{eq:ot_coup}
T_{a}^* = \argmin_{T}
\langle T, \tc \rangle,
\end{align} 
for $T \in \mathbb{R}^{N_{S} \times N_{\bar S}}$ $~\st~T\mathbf{1}_{\bar S} = \frac{1}{N_{S}}\mathbf{1}_{N_{S}}$ and $T^\top \mathbf{1}_{S} = \frac{1}{N_{\bar S}} \mathbf{1}_{\bar S}$, where $\mathbf{1}_{N_{S}}$ denotes a vector of ones of size $N_{S}$.
\item[2.] Perform a gradient update of the model parameters $\theta$ as 
\begin{align*}
\theta \leftarrow \theta - \epsilon_{\theta} \sum_{a \in \mathcal{A}}\sum_{i=1}^{N_{S}}\sum_{j=1}^{N_{\bar S}} T_{a}^*(i,j)\nabla_{g_{\theta}}\tc(s^i_a,\bar s^j)\nabla_{\theta}s^i_a.
\end{align*}
\end{itemize}
When $\mathcal{C}$ is an $L^p$ cost, it can be shown that $T_a^*$ is sparse. In this case the computational cost is ${\cal O}(N_{S} + N_{\bar S})$. 

The second method introduced in \citet{jiang2019wasserstein} (referred to as \emph{DPP}) is an approximate quantile post-processing method to match the model outputs to $p_{\bar S}$. 

The logistic regression (\emph{LR}) parameters $\theta^*$ used to initialize COT, DOT, and DPP were obtained using scikit-learn with default hyper-parameters \cite{pedregosa2011scikit-learn}.
The number of gradient updates for COT and DOT was set to $K=100,000$. 
An ablation study on the Adult dataset showed that COT gives similar performance for number of random Fourier features $D\geq 50$, for kernel variance in the range $\sigma^2=\{.01,.1,1\}$, and for the entropy and $L^2$ regularizations.
We only considered the summation for $i=j$ by imposing $N_{S}=N_{\bar S}$, which reduces the computation cost. 
Both COT and DOT required careful selection of the gradient update size $\epsilon_{\theta}$. In addition, COT required careful tuning of $\epsilon_{\theta}$ with the dual variables update size $\epsilon_{\lambda}$ and regularization strength $\lambda$. COT tended to display some instability during training, which could in some cases be alleviated with the constraint $\lambda_X(x) + \lambda_Y(x) = 0$.
\subsection{Fixed Level of Unfairness}
We compared LR, DOT, DPP, and COT on the following datasets from the UCI repository \cite{lichman2013uci}: 
\begin{description}
\setlength{\itemsep}{-1pt}  
\item[Adult Dataset.]
This dataset contains 14 attributes for 48,842 individuals.
The class label corresponds to annual income (below/above \$50,000). As sensitive attributes we considered race (Black and White) and gender (female and male), obtaining four groups.

\item[German Credit Dataset.]
This dataset contains 20 attributes for 1,000 individuals applying for loans. Each applicant is classified as a good or bad credit risk, \ie~as likely or not likely to repay the loan. As sensitive attributes we considered age (below/equal or above $30$ years old), obtaining two groups.
\item[Communities \& Crime Dataset.]
This dataset contains 135 attributes for 1994 communities. The class label corresponds to crime rate (below/above the 70-th percentile). As sensitive attributes we considered race (Black, White, Asian, and Hispanic) thresholded at the median, obtaining height groups. 
\end{description}
We used the Wasserstein-1 barycenter as target distribution $p_{\bar S}$ and the following metrics:
\begin{description}
\setlength{\itemsep}{1pt}  
\item[Err-.5:] Classification error at $\tau=0.5$ $\frac{1}{N}\sum_{n=1}^N\mathbbm{1}_{\hat y^n \neq y^n}$,
\item[Wass1:] Wasserstein-1 distance 
$\sum_{a \in \mathcal{A}}\W_1(p_{S_{a}}, p_{\bar S})$,
\item[SDD:] Strong Demographic Disparity\\ 
$~~~~~~\sum_{a \in \mathcal{A}}\mathbb{E}_{\tau\sim U[0,1]} |\mathbbm{P}(S_{a}>\tau) - \mathbbm{P}(S > \tau)|$,
\item[SPDD:] Strong Pairwise Demographic Disparity\\
$~~~~~~\frac{1}{2}\sum_{a,\bar a\in \mathcal{A}} \mathbb{E}_{\tau\sim U([0,1])} |\mathbbm{P}(S_{a}>\tau) - \mathbbm{P}(S_{\bar a} > \tau)|$,
\end{description}
computed as in \citet{jiang2019wasserstein}.

Test results are given in Table \ref{table:AGC}. We can see that the OT methods generally perform similarly, except for the German Credit dataset in which DOT has higher error for similar unfairness level. 
\begin{figure}[t]
\hspace*{-0.35cm}
\includegraphics[height=3.6cm,width=4.5cm]{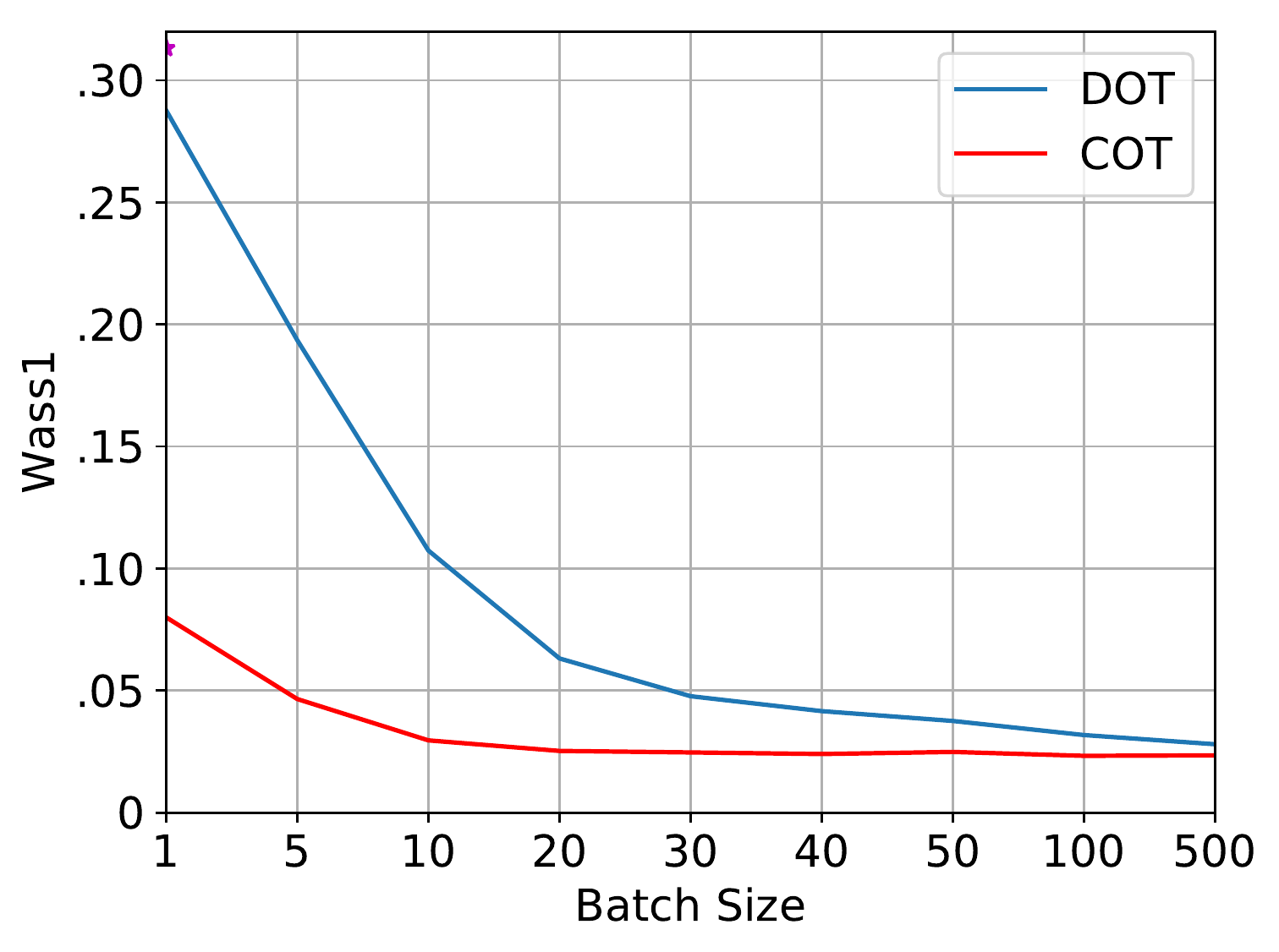}%
\hspace*{-0.1cm}
\includegraphics[height=3.6cm,width=4.5cm]{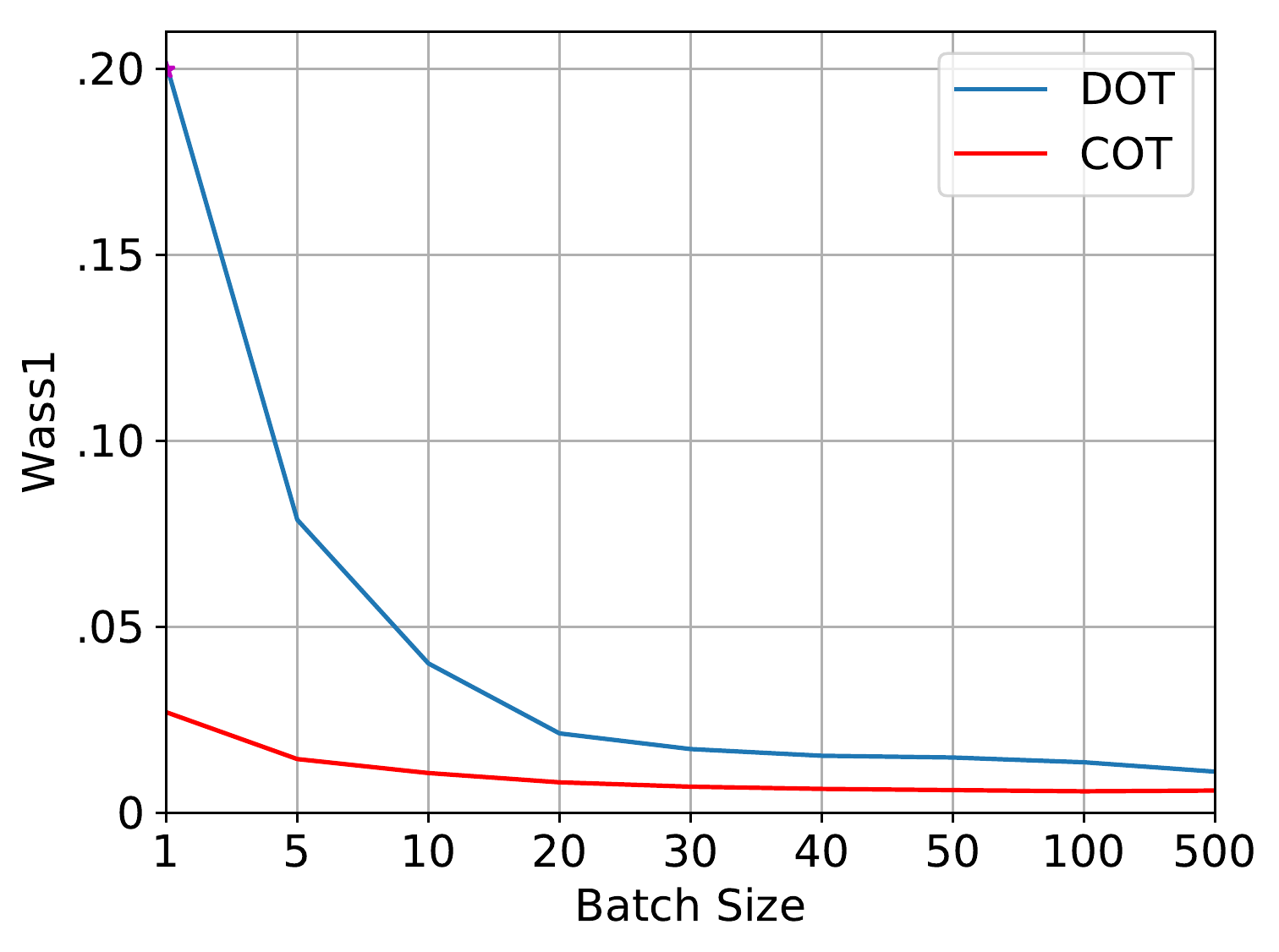}\\
\hspace*{-0.35cm}
\includegraphics[height=3.8cm,width=4.68cm]{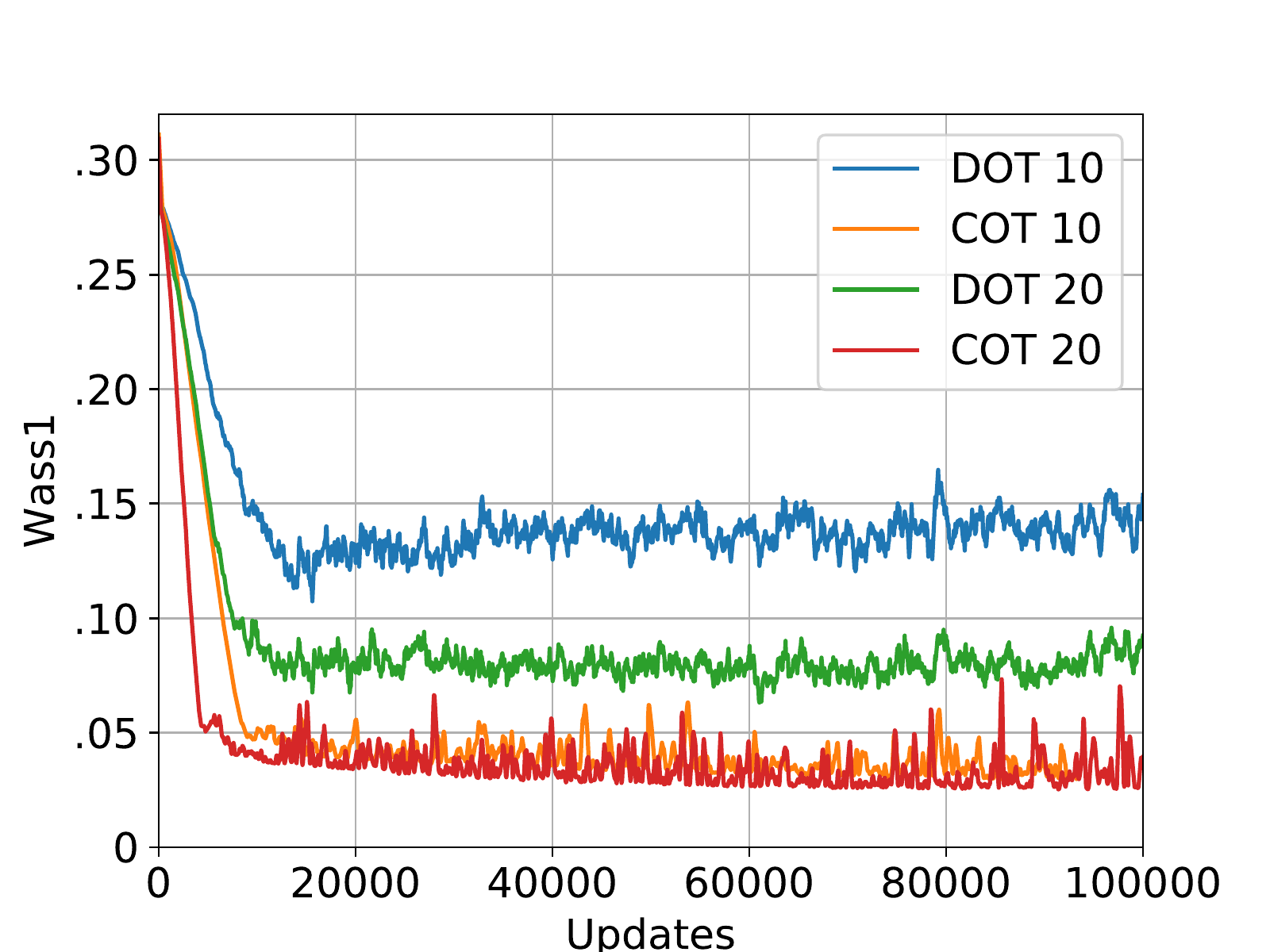}%
\hspace*{-0.22cm}
\includegraphics[height=3.8cm,width=4.68cm]{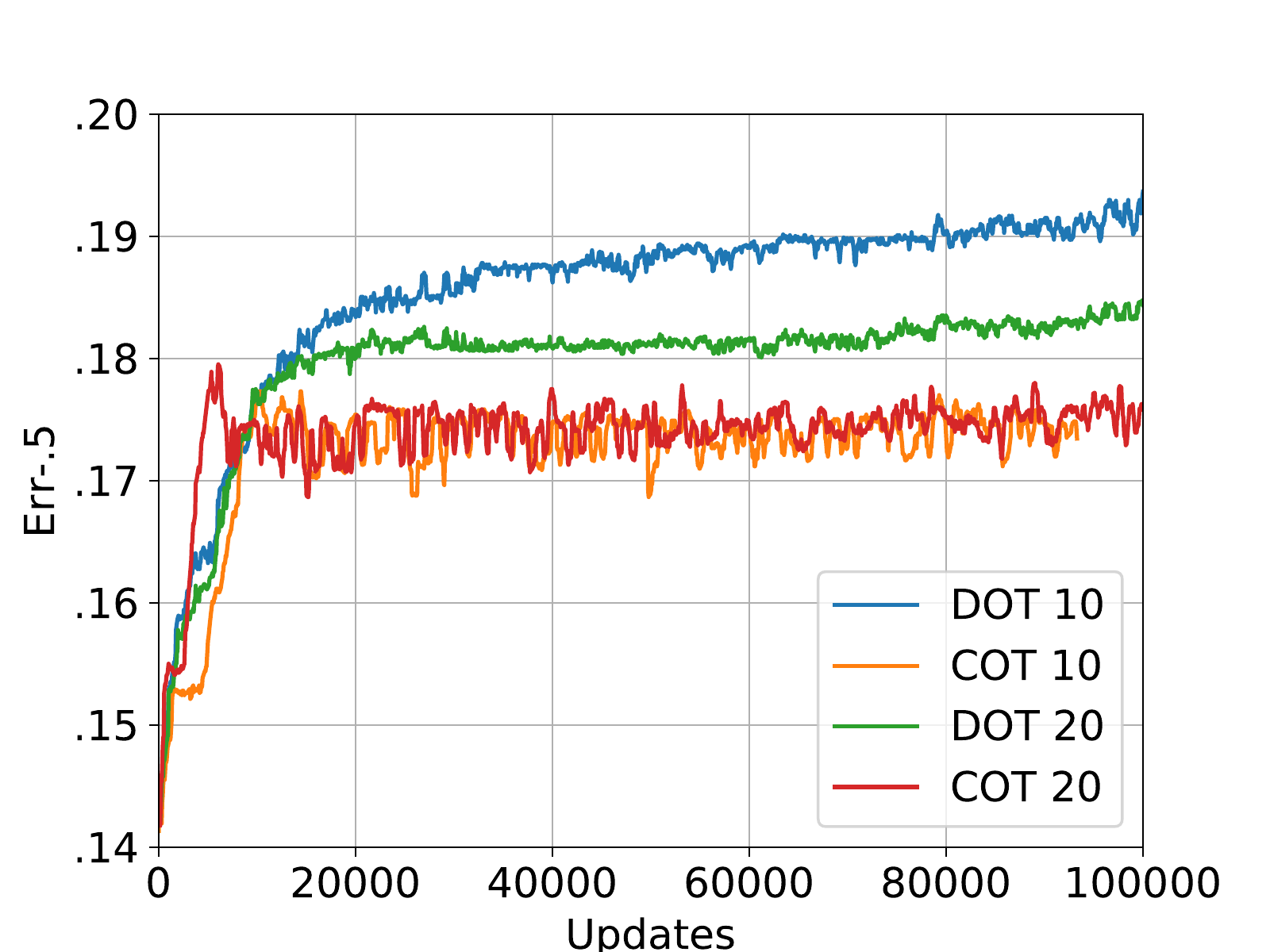}
\caption{{\bf Top Left:} Wass1 for DOT and COT on the Adult dataset using different batch sizes. The magenta star indicates Wass1 for LR. {\bf Bottom:} Evolution of Wass1 and Err-.5 for batch sizes 10 and 20 over the 100,000 gradient updates. {\bf Top Right:} The same as Top Left but considering only gender as sensitive attribute and the male outputs distribution as target distribution.}
\label{fig:BatchSizeDataSize}
\end{figure}
\paragraph{Solving the OT Problem with Limited Data.}
In real-world applications, the amount of data available to solve the OT problem might be limited---this could be the case, \eg, if data needs to be disregarded due to privacy issues. 

As the number of samples tends to infinity, the DOT Wasserstein-1 distance (and corresponding gradient) estimator approaches the true distance. Nevertheless, if the batch size is small, the resulting (random) estimator is biased as its expectation would not correspond to the true distance. For COT, we rely on the fact that it is possible to compute unbiased gradient estimators of the regularized Wasserstein-1 distance. This means that, using these estimators, it is possible to reach an arbitrarily (up to the batches variance) good estimator of the regularized Wasserstein-1 distance by sampling sufficiently many random batches. Nevertheless, this estimator will always be a biased version of the true distance, as it will converge to its regularized counterpart. 

To empirically test the effect of bias when little data is available, we compared COT and DOT on the Adult dataset with varying batch sizes $N_{S}=N_{\bar S}$. As we can see in \figref{fig:BatchSizeDataSize}(Top Left), for smaller batch sizes DOT has higher Wass1 than COT (the Wass1 value for LR is indicated with a magenta star). As we can see from the Wass1 and Err-.5 evolution over the 100,000 gradient updates for batch sizes 10 and 20 in \figref{fig:BatchSizeDataSize}(Bottom), DOT has higher error than COT despite higher Wass1. This suggests that the estimates of the optimal coupling matrix $T^*_a$ (Eq. (\ref{eq:ot_coup})) are inaccurate, as if the issue were \eg~with penalization or with a suboptimal choice of $\epsilon_{\theta}$ higher Wass1 would correspond to smaller error. 
\figref{fig:BatchSizeDataSize}(Top Right) shows that the same qualitative conclusions also hold for the simpler case of considering only gender as sensitive attribute and the male outputs distribution as target distribution.
\subsection{Changing Level of Unfairness}
\begin{figure}[t]
\centering
\includegraphics[height=3.4cm,width=7.8cm]{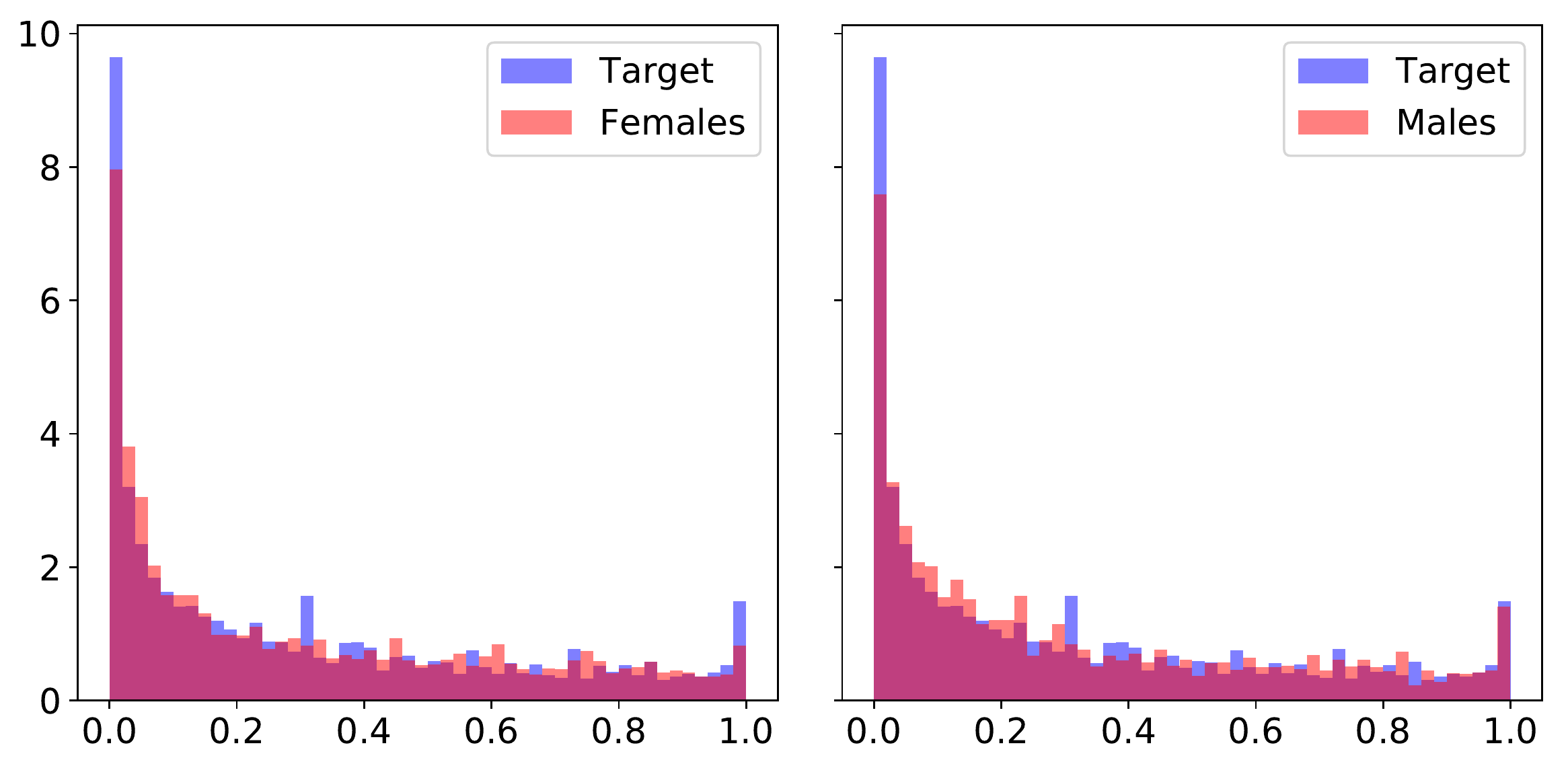}\\
\includegraphics[height=3.4cm,width=7.8cm]{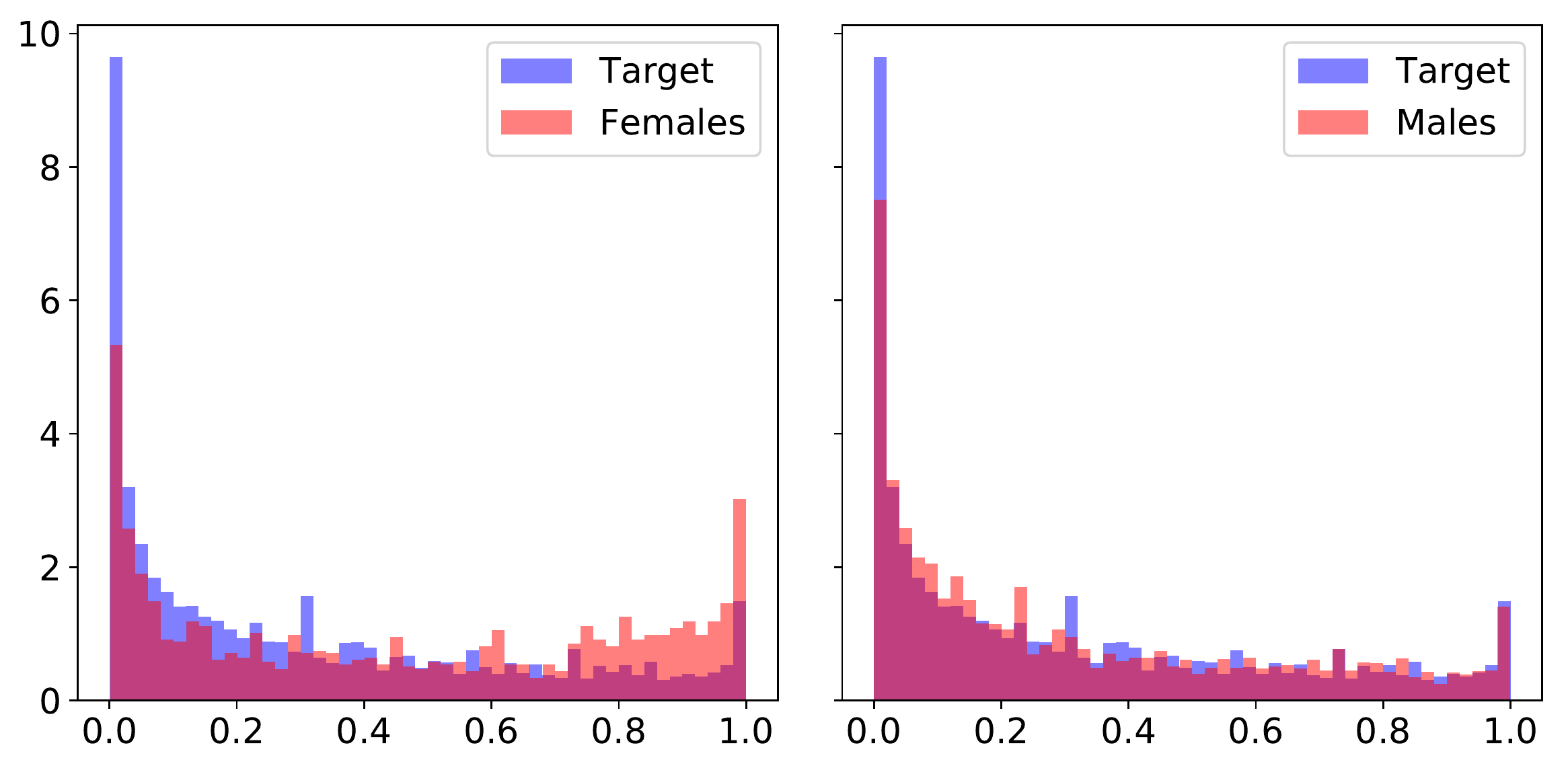}
\caption{Effect of change in unfairness level on COT on the Adult dataset, considering gender as sensitive attribute and the male outputs distribution as target distribution.
{\bf Top:} Histograms of the target distribution, and of the female and male outputs distributions produced by a model constrained to achieve SDP on a subset of the dataset with .1 and .3 probability of belonging to class 1 for female and male individuals respectively. {\bf Bottom:} Histograms produced by the same model, but using a subset of the dataset with .4 probability that a female individual belongs to class 1.}
\label{fig:TimeVarI}
\end{figure}
\begin{figure*}[t]
\hspace*{-0.3cm}
\subfloat[]{
\includegraphics[height=4cm,width=11cm]{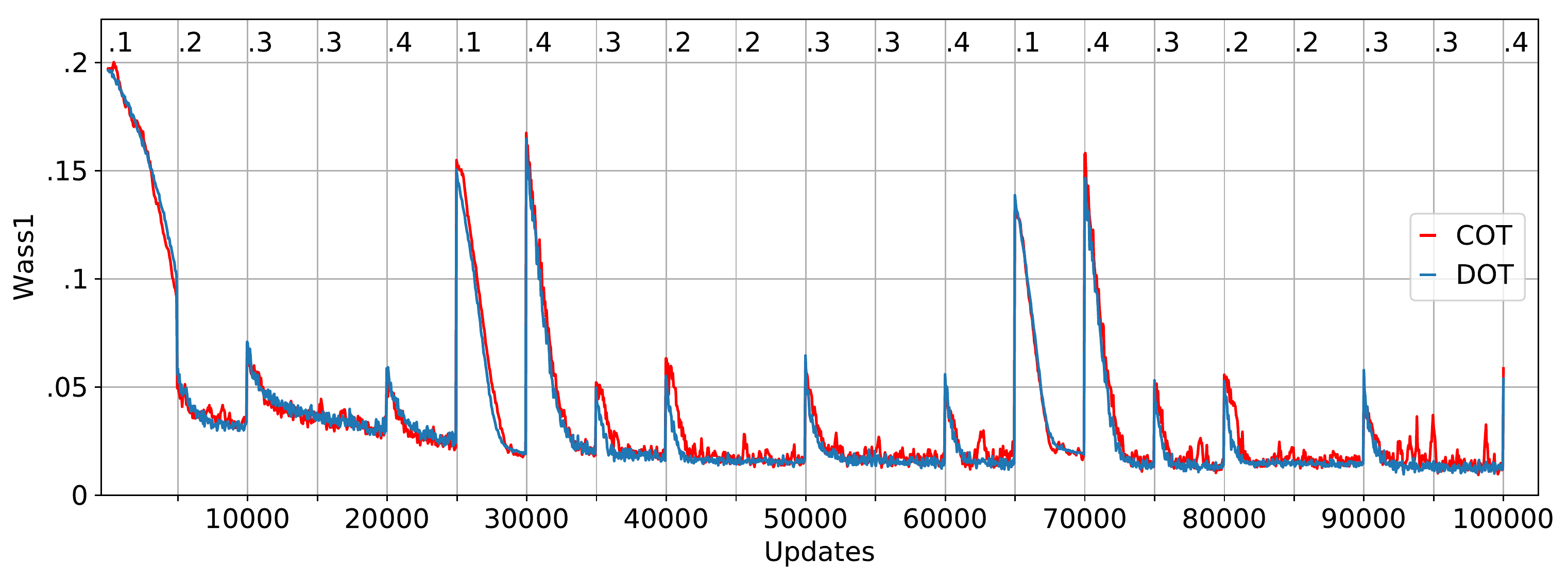}}
\hspace*{0.2cm}
\subfloat[]{
\raisebox{0.52cm}{\includegraphics[height=2.8cm,width=3.2cm]{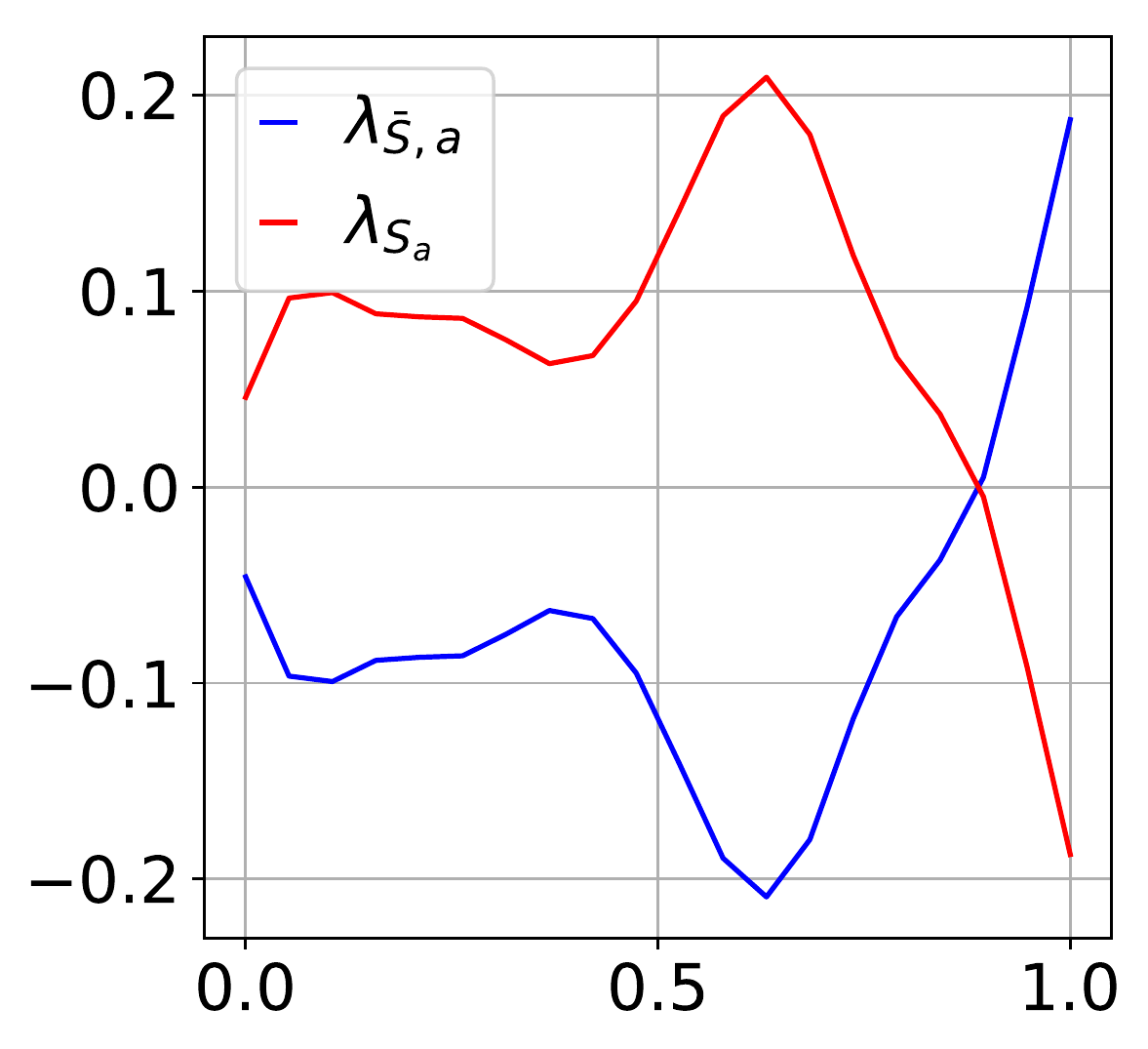}
\includegraphics[height=2.8cm,width=3.2cm]{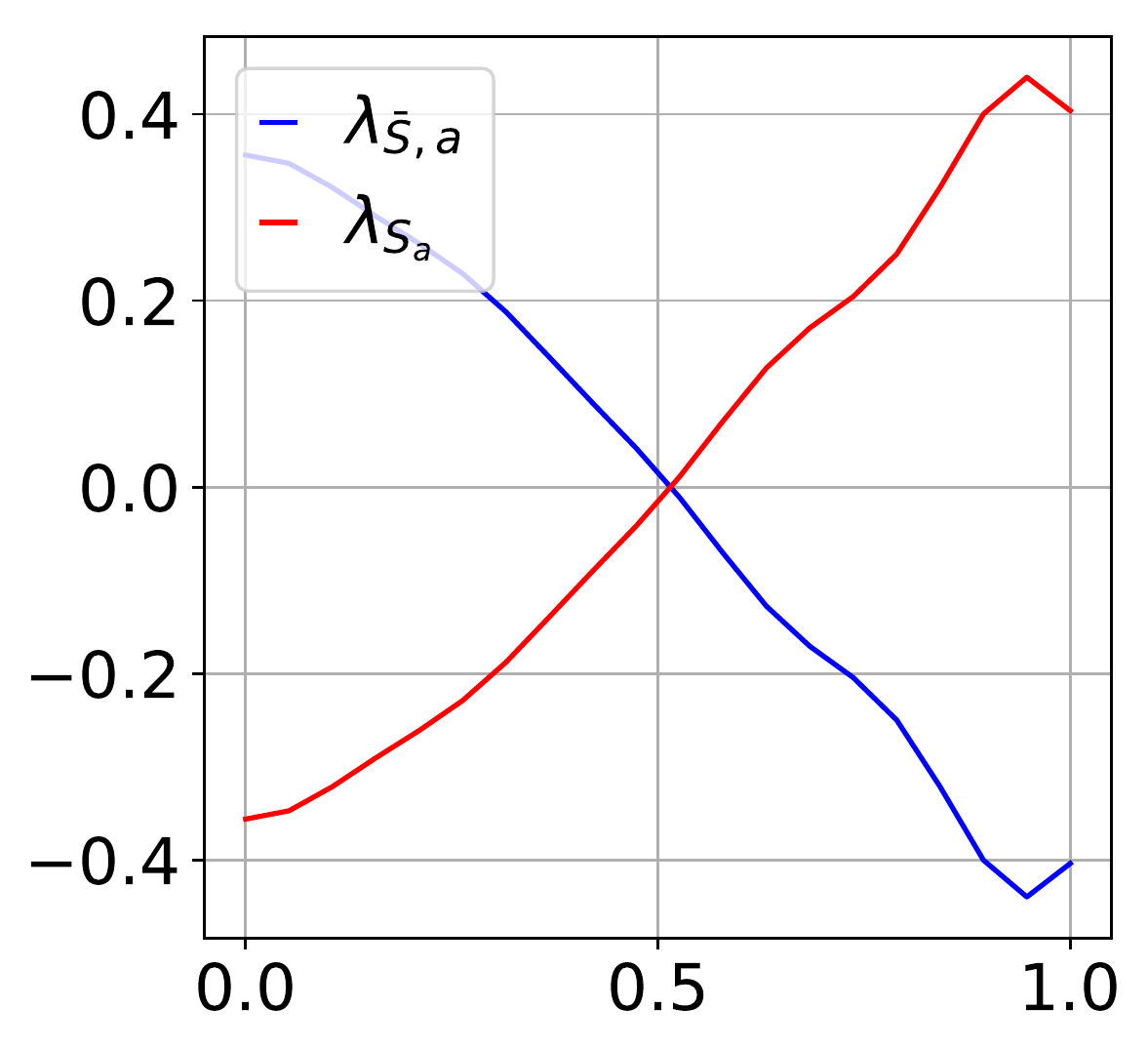}}
}
\caption{{\bf (a):} Wass1 for COT and DOT over 100,000 gradient updates, with changing (every 5,000 updates) probability of belonging to class 1 for female individuals as indicated at the top.
{\bf (b):} Dual variables for female individuals shortly before and after 30,000 updates in (a).}
\label{fig:TimeVarII}
\end{figure*}
In real-world applications, the data on which a system is deployed might contain a different level of unfairness than the data on which the system was enforced to be fair.
This could be the case, \eg, for a risk assessment instrument  deployed in a geographic area where over-policing in certain neighborhoods 
occurs at different rates than in the data used to train the system. Or for a system deployed later in the future, when biases have changed, \eg~as a consequence of the decisions 
taken by the system. 

In \figref{fig:TimeVarI}, we illustrate the effect of change in unfairness level on COT on the Adult dataset, considering gender as sensitive attribute and the male outputs distribution as target distribution. In the Adult dataset, the probability that a female individual belongs to class 1 (has annual income above \$50,000) is .1, as opposed to .3 for a male individual. 
\figref{fig:TimeVarI}(Top) shows histograms of the target distribution, $p_{\bS}$ (blue), and of the female and male outputs distributions, $p_{S_a}$ (red), produced by a model that was constrained to achieve SDP on a subset of the  dataset with the same probabilities of belonging to class 1. \figref{fig:TimeVarI}(Bottom) shows histograms produced by the same model, but using a subset of the dataset in which the probability of belonging to class 1 is .4 for female individuals and .3 for male individuals. Due to the increase of probability from .1 to .4, the model gives a distribution of female outputs that has too much mass toward higher probability of belonging to class 1. 

The underlying assumption when using SDP as fairness criterion is that existing dependencies between $A$ and $X$, and between $A$ and $Y$ are unfair (and therefore should not be encoded in the model output $S$). Therefore, a different relationship between $A$ and $X$ through a conditional distribution $p_t(X|A)$ that differs from the original distribution $p(X|A)$ would result in a different unfairness level.
A dataset ${\cal D}_t$ with underlying joint distribution $p_t(A,X,Y)=p(Y|A,X)p_t(X|A)p(A)$ can indirectly be obtained by sampling a subset of the original dataset ${\cal D}$ with a different probability of belonging to class 1, $p_t(Y=1|A)$. 

We created a sequence of such datasets, $\{{\cal D}_t\}$, by sampling subsets of the Adult dataset in which female individuals had the following probabilities of belonging to class 1: .2, ~.3, ~.3, ~.4, ~.1, ~.4, ~.3, ~.2, ~.2, ~.3, ~.3, ~.4, ~.1, ~.4, ~.3, ~.2, ~.2, ~.3, ~.3, ~.4. We trained COT and DOT for a total of 100,000 gradient updates starting from ${\cal D}_1\equiv {\cal D}$ and then moving to the next dataset in the sequence every 5,000 updates.

The resulting Wass1 is shown in \figref{fig:TimeVarII}(a) (values are plotted every 50 updates). Both COT and DOT adapt well to changes in unfairness levels and behave very similarly, although DOT is more stable. Interestingly, earlier in the training Wass1 is higher than later in the training when the probability of belonging to class 1 is also .1, and requires more updates to reach a low value. This indicates that, after the first modifications of the LR parameters $\theta^*$, $\theta$ reach a region of the space that enables fast adjustment. The fact that the minimum value of Wass1 decreases with progressing in the gradient updates also indicates that both COT and DOT modify the parameters in a way that is independent of the specific unfairness level contained in the dataset. 
Whilst on the Adult dataset it takes more than 50,000 updates for Wass1 to reach minimum levels given the logistic regression (LR) initialization, the right part of \figref{fig:TimeVarII}(a) shows that, on a subset of the dataset with the same probabilities of belonging to class 1, it does not take that many updates given an initialization at the model that was already adjusted for a different level of unfairness. Together with the left part, \figref{fig:TimeVarII}(a) shows that the parameters move to an area of the space that allows to quickly reach lower and lower Wass1.
Overall these experiments show that OT can quickly adjust to changes in the underlying unfairness level and without the need of retraining the model from scratch.

In \figref{fig:TimeVarII}(b), we show the dual variables for the female group shortly before and after gradient update 30,000 of \figref{fig:TimeVarII}(a). As we can see, before update 30,000 the values of $\lambda_{S_a}$ and $\lambda_{\bar S,a}$ are small and close to each-other over the interval [0, 1], indicating that the female and target distributions are similar. After update 30,000 the values are higher and differ mostly at the extremes of the interval [0,1], indicating  that the female and target distributions differ the most in this parts of the space. These two plots corresponds to the first histograms in \figref{fig:TimeVarI}(Top) and \figref{fig:TimeVarI}(Bottom) respectively.
\section{Conclusions}
We have proposed a stochastic-gradient fairness method based on a dual formulation of continuous OT that displays superior performance to discrete OT methods when little data is available to solve the OT problem, and similar performance otherwise. In addition, we have shown that both continuous and discrete OT methods are suitable to continual adjustments of the model parameters to adapt to changes in levels of unfairness that might occur in real-world applications of ML systems. 
\bibliography{COT}

\end{document}